\documentclass{article} 
\usepackage{iclr2023_conference,times}

\usepackage{amsmath,amsfonts,bm}

\def\eqref#1{equation~\ref{#1}}

\def\1{\bm{1}}

\DeclareMathAlphabet{\mathsfit}{\encodingdefault}{\sfdefault}{m}{sl}
\SetMathAlphabet{\mathsfit}{bold}{\encodingdefault}{\sfdefault}{bx}{n}

\usepackage{hyperref}
\usepackage{url}
\usepackage{graphicx}
\usepackage{float}

\title{Deep ensembles to improve uncertainty quantification of statistical downscaling models under climate change conditions}

\author{Jose González-Abad \& Jorge Baño-Medina\\
Instituto de Física de Cantabria (IFCA),\\  CSIC-
Universidad de Cantabria\\
Santander, Spain \\
\texttt{gonzabad@ifca.unican.es}}

\iclrfinalcopy 
\begin{document}

\maketitle

\begin{abstract}
Recently, deep learning has emerged as a promising tool for statistical downscaling, the set of methods for generating high-resolution climate fields from coarse low-resolution variables. Nevertheless, their ability to generalize to climate change conditions remains questionable, mainly due to the stationarity assumption. We propose deep ensembles as a simple method to improve the uncertainty quantification of statistical downscaling models. By better capturing uncertainty, statistical downscaling models allow for superior planning against extreme weather events, a source of various negative social and economic impacts. Since no observational future data exists, we rely on a pseudo reality experiment to assess the suitability of deep ensembles for quantifying the uncertainty of climate change projections. Deep ensembles allow for a better risk assessment, highly demanded by sectoral applications to tackle climate change.
\end{abstract}

\section{Introduction}

One of the most popular and extended tools for simulating the spatio-temporal evolution of climate systems are General Circulation Models (GCMs). Unfortunately, due to the complexity of the physical processes underlying climate dynamics, they usually incur in a high computational cost, resulting in coarse resolution climate fields. This hampers their applicability in the regional-to-local scale, where climate projections are crucial for some sectorial applications to elaborate adaptation and mitigation plans to climate change.

Statistical Downscaling (SD) \citep{maraun_statistical_2018} aims to generate high-resolution fields from coarse climate model outputs by learning a statistical relationship between a set of predictors (inputs) and predictands (outputs) representing the coarse and high resolution fields, respectively. Deep learning (DL) techniques \citep{goodfellow_deep_2016} have recently emerged as promising tools for downscaling coarse climate fields, showing excellent results in reproducing the observed local climate \citep{pan_improving_2019, bano_configuration_2020, sun_statistical_2021}.

For SD models to reliably compute high-resolution fields in future scenarios, it is crucial that they generalize to unseen conditions (stationary assumption), such as those of climate change. Most DL models adapted to the SD task minimize losses such as the mean squared error, which leads models to fit the mean of the training data, no providing information about the confidence in their predictions. Taking this uncertainty into account is extremely important when applying these models to future scenarios influenced by climate change, where the stationarity assumption may be violated. Also, being able to capture extremes is key for planning against extreme weather events (e.g., droughts, heat waves, extreme heat) whose frequency is expected to grow due to climate change. By improving the modeling of the uncertainty of SD models we can better capture these events when downscaling climate fields under climate change conditions. 

To quantify uncertainty in SD, in \citep{bano_configuration_2020} authors model the parameters of a distribution over each gridpoint of the downscaled field, relying on information about the climate variable to choose the appropriate distribution (e.g., Gaussian for temperature). Subsequent work \citep{quesada_repeatable_2022} follows the same procedure as a way to improve the distributional properties of the predictions. In \citep{vandal_quantifying_2018} authors rely on Monte Carlo Dropout \citep{gal_dropout_2016}, an approximation to Bayesian inference through the use of dropout at test time, to quantify uncertainty in the downscaling of precipitation, obtaining acceptable results. Recently, several works explore generative modeling for the statistical downscaling task \citep{price2022increasing, harris_generative_2022, accarino_msg_2021}, obtaining benefits from its stochastic properties. Among these works, only \citep{bano_downscaling_2022} quantifies the uncertainty in climate change projections.

A simpler way to achieve uncertainty quantification is through deep ensembles. It consists of training several models in parallel on the same data and then combining the predictions through an aggregation operation (e.g. mean), which generally improves performance compared to a single model \citep{dietterich_ensemble_2000}. In \citep{lakshminarayanan_simple_2017} authors show how this technique also allows for a better predictive uncertainty estimates, however, its application to SD has not yet been studied. In this work we explore a natural extension of \citep{bano_configuration_2020} through the use of deep ensembles, a simple technique that does not require tuning hyperparameters or modifying the model topology. This leads to a better quantification of uncertainty of climate change projections in future scenarios.

\section{Experimental Framework}

Due to the lack of observations in future periods, we rely on a pseudo reality experiment to evaluate the effectiveness of deep ensembles for SD. This experiment is necessary if we need to reveal whether the relationship learned in a historical period is able to extrapolate to future scenarios \citep{maraun_statistical_2018}. We use the regional climate model (RCM) CanRCM4 \citep{scinocca_coordinated_2016} as pseudo observations (predictand) and its driving GCM (CanESM2) \citep{chylek_observed_2011} as predictor. This RCM is part of the
Coordinated Regional Climate Downscaling Experiment (CORDEX\footnote{https://cordex.org/}) initiative, which provides multi-model ensembles of RCMs at the continental level. In addition, the chosen RCM rely on spectral nudging \citep{von_spectral_2000}, a technique used to improve the day-to-day correspondence of dynamical downscaling models \citep{maraun_statistical_2018}. 

We rely on the DeepESD convolutional model introduced in \citep{bano_configuration_2020, bano_downscaling_2022} to downscale temperature, which has demonstrated the ability to accurately reproduce the local climate. It is composed of three convolutional layers with 50, 25 and 10 kernels (with ReLU activations). The output of the last convolutional layer is flattened and passed to a dense linear layer to compute the $\mu$ and $\sigma$ vectors, containing the parameters of the corresponding Gaussian distribution for each gridpoint of the predictand. As a result, the neural network defines a mapping $[\mu, \sigma] = f^w(x)$ (parameterized by $w$), where $x$ is the input of the model, in our case the CanESM2 predictors.

We follow previous work \citep{bano_configuration_2020} and select as predictors 3 large-scale variables (geopotential height, air temperature and specific humidity) at 4 different vertical levels (500, 750, 800 and 1000 hPa) from CanESM2 (2.8125$^\circ$ resolution). These variables have been shown to be adequate drivers of local temperature \citep{maraun_statistical_2019, gutierrez_reassessing_2013, gonzalez-abad_interp_2022}. For the target predictand, we lean on the near-surface air temperature of CanRCM4 (0.22$^\circ$ resolution), which justifies modeling the parameters of a Gaussian distribution. As region to perform the experiment, we select a western area of United States (25$^\circ$ to 55$^\circ$ in latitude and -135$^\circ$ to -100$^\circ$ in longitude).

The DeepESD model is fitted in a training set spanned by the years 1980-2002 by minimizing the negative log likelihood (NLL) of a Gaussian distribution at each predictand gridpoint. To avoid overfitting, we follow an early stopping strategy on a random 10\% split of the training set. Following previous work \citep{bano_suitability_2021, bano_downscaling_2022}, the evaluation of these models is performed in three different future periods (2006-2040, 2041-2070 and 2071-2100). Predictors are scaled at gridpoint level, taking as reference the mean and standard deviation computed over the training set.

For the ensemble, we train 10 different models in parallel, following the procedure described in the previous paragraph. To aggregate the models we follow \citep{lakshminarayanan_simple_2017}, so the mean and variance of the final ensemble are given by
\begin{equation}
\mu_*(x) = \frac{1}{M} \sum_{m} \mu_{\theta_m}(x) \quad \quad
\sigma^{2}_*(x) = \frac{1}{M} \sum_{m} (\sigma^{2}_m(x) + \mu^{2}_m(x)) - \mu^{2}_*(x)
\end{equation}
where $M$ is the number of models composing the ensemble and $\mu_{\theta_m}(x)$ and $\sigma^{2}_m(x)$ are the mean and variance of the $m$ ensemble member.

\section{Results}

\begin{figure}[h]
  \centering
  \includegraphics[scale=0.38]{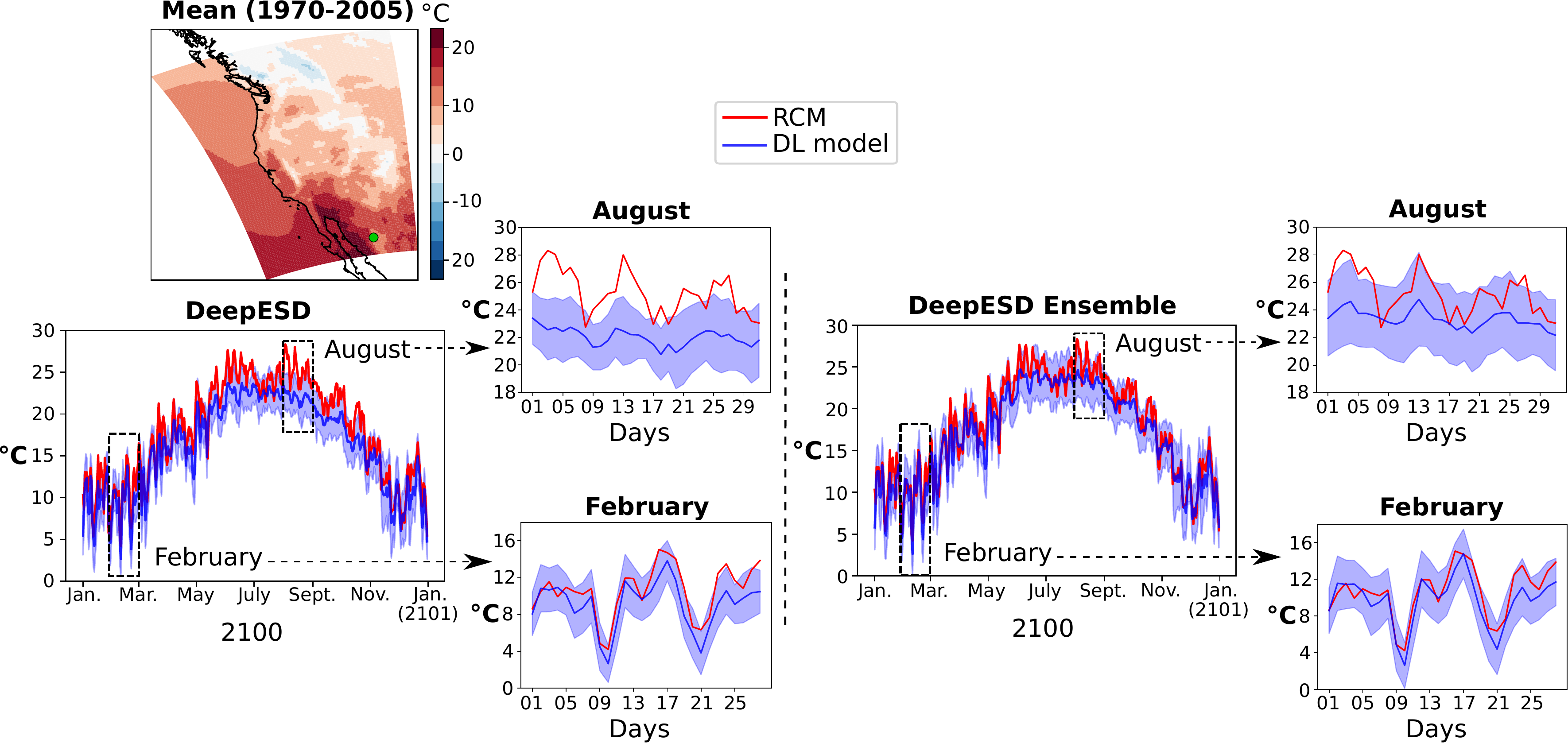}
  \caption{Mean and 95\% confidence intervals of the DeepESD and DeepESD Ensemble models for the predicted near-surface air temperature (in blue) of the year 2100. These are compared with the corresponding RCM target values (in red). In addition, for each model, an enlarged plot of the months of August and February is shown. The climatology of the historical period (1970-2005) for the RCM is also shown (upper-left corner). This map shows the gridpoint to which the time series belongs (green point).}
  \label{fig:1}
\end{figure}

Figure \ref{fig:1} shows the RCM target values (in red), as well as the prediction (in blue) from DeepESD and the DeepESD Ensemble for the year 2100. For the prediction, the mean and the 95\% confidence interval are plotted. The displayed time series correspond to the gridpoint indicated with a green point in the upper-left figure, which represents the climatology of the RCM in the historical period. For each of the models, a time series of the August and February months is also depicted. As long as the RCM target value falls within the interval confidence it means that the model is able to account for the uncertainty of that given value, otherwise it fails to do so. As we can see in the climatology plot, the selected point falls into an area with warm temperatures. Overall, DeepESD model fits RCM temperature accurately, however for the warmer months (June, July and August) we can observe how the mean is biased toward lower values. If we turn our attention to August, we can observe how, besides a biased mean, the confidence interval is not able to capture the RCM values. This is not the case for winter months (e.g. February), where we see how the mean is not biased, and the confidence interval accurately captures the RCM values. The DeepESD Ensemble models does not suffer from this bias in the warmer months, moreover, we can see how in August the confidence interval is able to capture most of the values. For February, the performance of DeepESD Ensemble is similar to that of DeepESD.

\begin{figure}[h]
  \centering
  \includegraphics[scale=0.4]{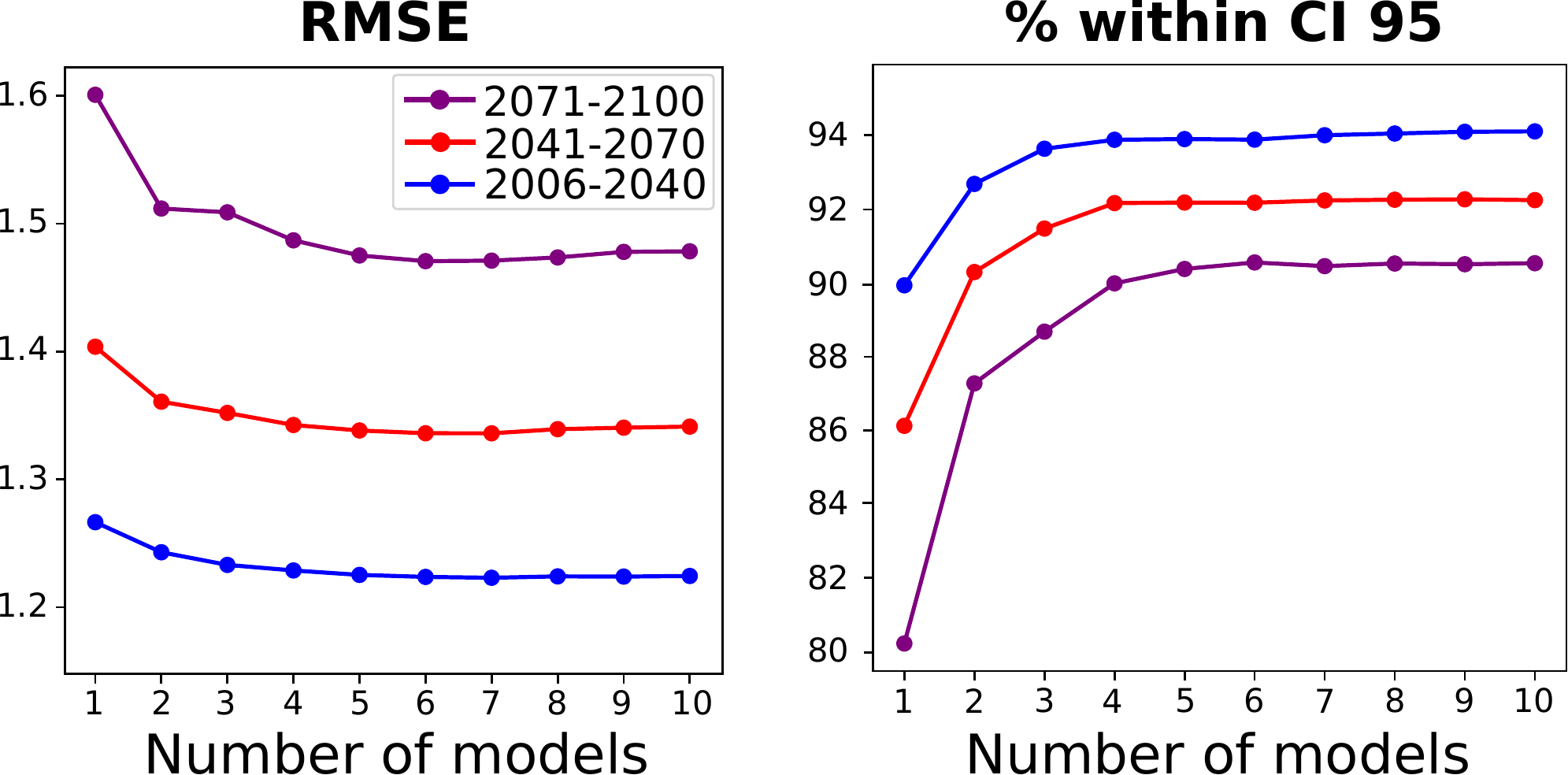}
  \caption{Spatial mean of RMSE and ratio of target samples within the predicted 95\% confidence interval for summer days for the three intervals considered (2006-2040, 2041-2070 and 2071-2100), with respect to the number of models composing the ensemble.}
  \label{fig:2}
\end{figure}

To better understand the advantages of deep ensembles in both mean and uncertainty quantification, we display in Figure \ref{fig:2} the spatially averaged Root Mean Square Error (RMSE) and the ratio of the target RCM values falling within the 95\% confidence interval for the three future periods (2006-2040, 2041-2070 and 2071-2100). We focus only on summer days, since it is on these days that DeepESD and DeepESD Ensemble disagree the most (see Figure \ref{fig:1}). RMSE values are higher as we move forward in time. The improvement caused by adding models to the ensemble stops around 3-4 models. This improvement is higher as we move forward in time, from almost no improvement in the 2006-2040 period to a 0.1$^\circ$ improvement for 2071-2100. A similar behaviour is observed for the ratios of target samples within the 95 \% confidence interval. In this case, the improvement also gets larger as we move forward in time. For this metric, DeepESD Ensemble benefits from a higher number of models in the ensemble (around 5-6).

\begin{figure}[h]
  \centering
  \includegraphics[scale=0.27]{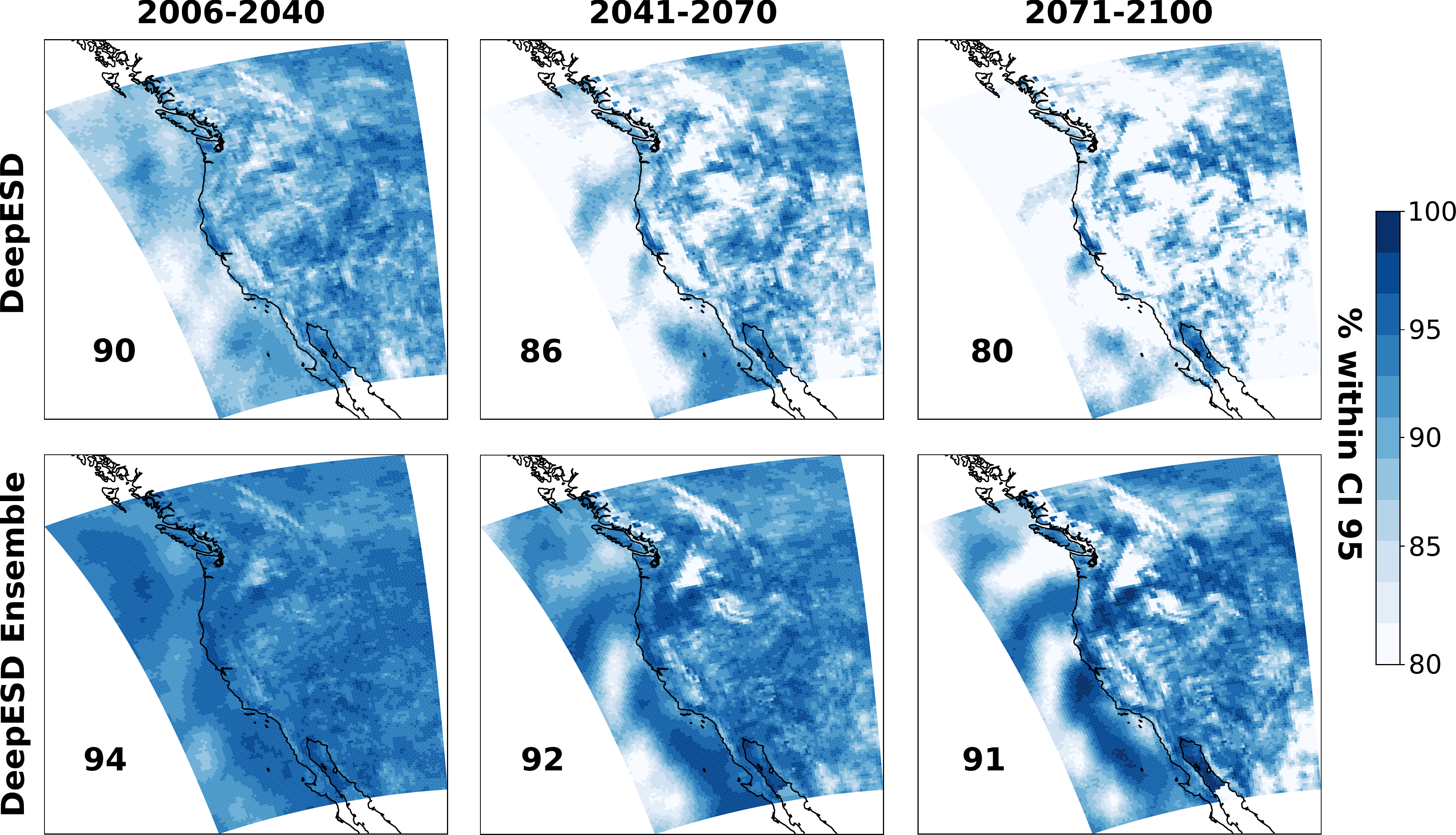}
  \caption{Ratio of RCM target values within the predicted 95\% confidence interval for summer days for the three intervals considered (2006-2040, 2041-2070 and 2071-2100) in columns, for the DeepESD and DeepESD Ensemble models in rows. The number in each panel represents the spatial mean.}
  \label{fig:3}
\end{figure}

Figure \ref{fig:3} shows the same ratio over the summer days but without spatially averaging, that is the value for each gridpoint of the predictand. These are displayed for the two models (DeepESD and DeepESD Ensemble) in rows and for the three different intervals (2006-2040, 2041-2070 and 2071-2100), in columns. We can see how the difference between the two models is remarkable, especially as we move forward in time, going from an average 4\% improvement to an 11\%. Most of the improvements occur at land-based gridpoints. Some areas of the North Pacific Ocean still do no benefit from the deep ensembles.

\section{Conclusion}

In this work we have extended previous work on SD by focusing on a better uncertainty quantification in order to capture the uncertainty of climate change projections. To achieve this, we rely on deep ensembles, a technique that showed improved performance in the quantification of uncertainty. For the first time we study the effect of this simple technique in SD, and find that it provides improvements in the predicted confidence interval, especially in future scenarios with unseen conditions.

Both models are able to model the uncertainty for the cold months, however for the warmer months (summer) DeepESD is biased, underestimating and therefore failing to capture much of these values. However, the DeepESD Ensemble manages to fit the mean of the RCM, capturing a large part of the positive RCM values due to its improved confidence interval. This becomes clearer when studying the RMSE and the ratio on summer days. We see how the ensemble brings considerable advantages, especially when we move forward in time, where we expect to find an upward trend in temperature due to climate change and therefore warmer extremes. 

In climate change scenarios, quantifying uncertainty is extremely important since it helps us to better capture extreme weather events, especially under climate change conditions, where their intensity and frequency are expected to worsen. Deep ensembles appear to be a natural and straightforward extension of current methods for modeling uncertainty, thus helping SD to better tackle climate change. In this work we validate this technique in a pseudo reality experiment, in future work we plan to apply it directly to SD and study its benefits in a real-case scenario. Additionally, we plan to extend this study to other climate fields, such as precipitation.

\subsubsection*{Acknowledgments}
J. González-Abad would like to acknowledge the support of the funding from the Spanish Agencia Estatal de Investigación through the Unidad de Excelencia María de Maeztu with reference MDM-2017-0765. J. Baño-Medina acknowledges support from Universidad de Cantabria and Consejería de Universidades, Igualdad, Cultura y Deporte del Gobierno de Cantabria via the ``instrumentación y ciencia de datos para sondear la naturaleza del universo'' project.

\bibliography{iclr2023_conference}
\bibliographystyle{iclr2023_conference}


\end{document}